\documentclass[conference]{IEEEtran}
\IEEEoverridecommandlockouts
\usepackage{cite}
\usepackage{amsmath,amssymb,amsfonts}
\usepackage{algorithmic}
\usepackage{graphicx}
\usepackage{textcomp}
\usepackage{xcolor}
\usepackage{flushend}
\usepackage{subcaption}
\usepackage{xurl}
\newcommand{\etal}{\textit{et al}. }
\newcommand{\ie}{\textit{i}.\textit{e}., }

\def\BibTeX{{\rm B\kern-.05em{\sc i\kern-.025em b}\kern-.08em
    T\kern-.1667em\lower.7ex\hbox{E}\kern-.125emX}}

\begin{document}

\title{Understanding Divergent Framing of the Supreme Court Controversies: Social Media vs. News Outlets \\
}
\makeatletter
\newcommand{\linebreakand}{%
  \end{@IEEEauthorhalign}
  \hfill\mbox{}\par
  \mbox{}\hfill\begin{@IEEEauthorhalign}
}
\makeatother

\author{\IEEEauthorblockN{Jinsheng Pan}
\IEEEauthorblockA{\textit{Department of Computer Science} \\
\textit{University of Rochester}\\
Rochester, USA \\
jpan24@ur.rochester.edu}
\and
\IEEEauthorblockN{Zichen Wang\textsuperscript{*} \thanks{\textsuperscript{*} Equal Contribution.}}
\IEEEauthorblockA{\textit{Department of Computer Science} \\
\textit{University of Rochester}\\
Rochester, USA \\
zwang189@ur.rochester.edu}
\and
\IEEEauthorblockN{Weihong Qi\textsuperscript{*}}
\IEEEauthorblockA{\textit{Department of Political Science} \\
\textit{University of Rochester}\\
Rochester, USA \\
wqi3@ur.rochester.edu}
\linebreakand
\IEEEauthorblockN{Hanjia Lyu}
\IEEEauthorblockA{\textit{Department of Computer Science} \\
\textit{University of Rochester}\\
Rochester, USA \\
hlyu5@ur.rochester.edu}
\and
\IEEEauthorblockN{Jiebo Luo}
\IEEEauthorblockA{\textit{Department of Computer Science} \\
\textit{University of Rochester}\\
Rochester, USA \\
jluo@cs.rochester.edu}
}

\maketitle

\begin{abstract}
Understanding the framing of political issues is of paramount importance as it significantly shapes how individuals perceive, interpret, and engage with these matters. While prior research has independently explored framing within news media and by social media users, there remains a notable gap in our comprehension of the disparities in framing political issues between these two distinct groups. To address this gap, we conduct a comprehensive investigation, focusing on the nuanced distinctions both qualitatively and quantitatively in the framing of social media and traditional media outlets concerning a series of American Supreme Court rulings on affirmative action, student loans, and abortion rights. Our findings reveal that, while some overlap in framing exists between social media and traditional media outlets, substantial differences emerge both across various topics and within specific framing categories. Compared to traditional news media, social media platforms tend to present more polarized stances across all framing categories. Further, we observe significant polarization in the news media's treatment (\ie \texttt{Left} vs. \texttt{Right} leaning media) of affirmative action and abortion rights, whereas the topic of student loans tends to exhibit a greater degree of consensus. The disparities in framing between traditional and social media platforms carry significant implications for the formation of public opinion, policy decision-making, and the broader political landscape.
\end{abstract}

\begin{IEEEkeywords}
framing analysis, public opinion, policy perception
\end{IEEEkeywords}

\section{Introduction}
Understanding the framing of political issues is crucial because it influences how individuals perceive, interpret, and engage with those issues~\cite{rothman1997shaping}. To frame is to ``select some aspects of a perceived reality and make them more salient in a communicating text, in such a way to promote a particular problem definition, causal interpretation, moral evaluation, and/or treatment recommendation for the item described"~\cite{entman1993framing}. More specifically, framing can contribute to political polarization~\cite{clifford2019emotional}. Different news outlets might present the same political issue in contrasting ways, reinforcing existing beliefs and deepening divisions~\cite{panbias}. Framing can cast policies in a more favorable or unfavorable light, potentially affecting policy support and voting decisions. While previous research has independently examined framing by the news media~\cite{valenzuela2017behavioral} and social media users~\cite{mendelsohn2021modeling}, there remains a gap in our understanding regarding the distinctions in framing political issues between these two groups.

The American Supreme Court's term, ending on June 30, 2023, was marked by a series of rulings on cases concerning affirmative actions~\cite{hurley2023supreme}, student loans~\cite{liptak2023supreme}, and gay rights~\cite{chung2023us}. These decisions garnered substantial attention from news media outlets but also ignited fervent debates across various social media platforms. This scenario offers a valuable prospect to explore the characterization of the framing surrounding these political matters and to scrutinize the shared aspects and disparities in framing as presented by both news media and social media users on a large scale.

\begin{table*}[t]
\centering
\caption{Query keywords for news title collection. Note that the keywords are converted to lowercase to form search queries. }
\label{tab:keywords_GNews} 
\begin{tabular}{ccc}
\hline
Affirmative Action & Student Loan & Abortion Rights\\
\hline 
affirmative action&student loan &abortion supreme court  \\
 Students for Fair Admissions v. Harvard & student debut harvard admission & Dobbs v.Jackson Women's Health Organization  \\
harvard admission &Biden v. Nebraska & \\
\hline 
\end{tabular}
\end{table*}

Our study focuses on the rulings of the American Supreme Court in three pivotal cases: \textit{Students for Fair Admissions v. Harvard}, \textit{Biden v. Nebraska}, and \textit{Dobbs v. Jackson Women's Health Organization}, regarding affirmative action, student loan, and abortion rights, respectively. We pay attention to these three cases because they have a wide social impact and highlight the disparities between public and media outlets.

To investigate the nuances of framing across various platforms, our research seeks answers to the following questions:
\begin{itemize}
\item RQ1: Do framing differences exist between news media and social media?
\item RQ2: Are framing differences discernible between supporters and opponents across the three cases?
\end{itemize}

In our study, we collect 4,784 news headlines and 41,143 Reddit comments. We then detect the frame and stance for each Reddit comment or news title. Our findings indicate that while there is a convergence on certain frames, significant differences exist across topics when comparing social media platforms to traditional news outlets. Furthermore, social media platforms display a broader range of stances compared to traditional media, suggesting heightened polarization on these platforms. Notably, both platforms concur on the \textit{Crime and Punishment} frame in relation to abortion rights, reflecting a consensus among the broader public. Our observations also highlight polarization in news media stances on affirmative action and abortion rights, whereas the student loan topic tends to maintain more consensus.

\begin{table}[t]
\caption{Summary statistics of the collected Reddit comments and news titles.} 
\label{tab:summary} 
\centering
\begin{tabular}{l*{3}{c}r}
\hline 
Topic & \# of news titles & \# of Reddit comments \\
\hline
Affirmative action & 1,096 & 5,577 \\
Student loan &	1,016 &	16,174 \\
Abortion Rights &	2,672 &	19,392 \\
\hline
Total	& 4,784 & 41,143 \\
\hline
\end{tabular}
\end{table}

\begin{table*}[t]
\caption{Description of each frame. Our analysis incorporates 14 frames defined by Mendelsohn \etal~\cite{mendelsohn2021modeling}, encompassing the economic, political, cultural, ethical, health, and legal dimensions of controversial issues. } 
\label{tab:frame}  
\resizebox{2\columnwidth}{!}{%
\begin{tabular}{l{l}}
\hline
Frame            & Description  \\
\hline
Thematic & Message is more abstract, placing stories in broader political and social contexts. \\
Episodic & Message provides concrete information about specific people, places, or events. \\
Economics & Financial implications of an issue.\\
Capacity \& Resources  &	The availability or lack of time, physical, human, or financial resources.  \\
Morality \& Ethics &	Perspectives compelled by religion or secular sense of ethics or social responsibility. \\
Fairness \& Equality &  The (in)equality with which laws, punishments, rewards, and resources are distributed. \\
Legality, Constitutionality
\& Jurisdiction	& Court cases and existing laws that regulate policies; constitutional interpretation
legal processes; jurisdiction. \\
Crime \& Punishment & The violation of policies in practice and the consequences of those violations. \\
Security \& Defense & Any threat to a person, group, or nation and defenses taken to avoid that threat. \\
Health \& Safety & Health and safety outcomes of a policy issue, discussions of health care. \\
Quality of Life & Effects on people’s wealth, mobility, daily routines, community life, happiness, etc. \\
Cultural Identity & Social norms, trends, values, and customs integration/assimilation efforts. \\
Public Sentiment & General social attitudes, protests, polling, interest groups, public passage of laws. \\
Political Factors \& Implications &
Focus on politicians, political parties, governing bodies, political campaigns
and debates; discussions of elections and voting. \\
Policy Prescription &
Evaluation Discussions of existing or proposed policies and their effectiveness \\
External Regulation. \&
Reputation &
Relations between nations or states/provinces; agreements between governments;
perceptions of one nation/state by another. \\
\hline

\end{tabular}%
}
\end{table*}

\section{Related Work}

News discourse has been found to be carefully crafted in the American political processes. This phenomenon is fueled, in part, by the increasing proactive efforts of politicians and interest groups to magnify their interpretations of the essence of a given issue~\cite{pan1993framing}. Framing analysis within this context has garnered significant attention in scholarly research~\cite{an2009news, lawlor2017deciding,barry2011news, mcginty2014news}. For instance, while Canada is frequently regarded as a stronghold of encouragement for immigrants and refugees, Lawlor and Tolley~\cite{lawlor2017deciding} revealed a nuanced landscape of support. Canadian sentiments exhibit differentiation between economic immigrants and those who enter the country on humanitarian grounds. Economic immigrants are often cast within an economic framework, while heightened attention centers on the legitimacy of refugee claims, potential security concerns, and the perceived impact of refugees on social welfare programs. Recent research has also delved into the interplay between media frames and public opinion, examining how frames influence individuals' attitudes and beliefs~\cite{de2011direct, giles2009psychology, schnell2001assessing, lim2009frame}. Take the membership of Turkey in the European Union as an example, through the presentation of positively and negatively valenced news frames to two distinct groups of participants, De Vreeze \etal~\cite{de2011direct} uncovered a noteworthy disparity in the degree of support for Turkish membership.

Additionally, digital media's rise has led to large-scale investigations into how framing operates across various online platforms, and how audiences interact with and share framed content~\cite{van2013public, mendelsohn2021modeling, siapera2018refugees, heidenreich2019media}. By analyzing the word frequency in social media posts and media coverage, Van der Meer and Verhoeven~\cite{van2013public} conducted a study to examine the difference in framing surrounding the explosion of the Dutch Chemie-Pack plant in Moerdijk. Their research investigated how news media and the public on social media framed this event. Their findings highlighted that the accessibility of social media, facilitated rapid and widespread self-communication among the masses. Consequently, a swift emergence of public crisis framing, based on conjectures, occurred in the absence of immediate news media coverage. However, as news media reports became available, the public's apprehension diminished, aligning their crisis perception with the framing presented by the media. To facilitate large-scale framing analysis, Boydstun \etal~\cite{boydstun2013identifying} developed a list of issue-generic policy frame labels for articles spanning various topics, encompassing aspects such as economy, morality, fairness, and so on. Iyengar~\cite{iyengar1994anyone} introduced narrative framing which is classified as either episodic or thematic. Building on these, Mendelsohn \etal~\cite{mendelsohn2021modeling} curated a dataset tailored for framing analysis.

Our research augments the existing literature by introducing a cross-platform framing analysis, focusing particularly on prevalent controversial issues within the U.S. Our exploration of framing disparities offers a novel perspective on the interplay between public opinions and framing in news outlets. This deepens insights beyond the findings of current studies~\cite{de2011direct, giles2009psychology, schnell2001assessing, lim2009frame}, and broadens our comprehension of the discrepancies between digital media platforms and traditional news sources.

\begin{figure*}[t]
\centering
\begin{subfigure}{0.45\textwidth}
    \includegraphics[width=0.9\textwidth]{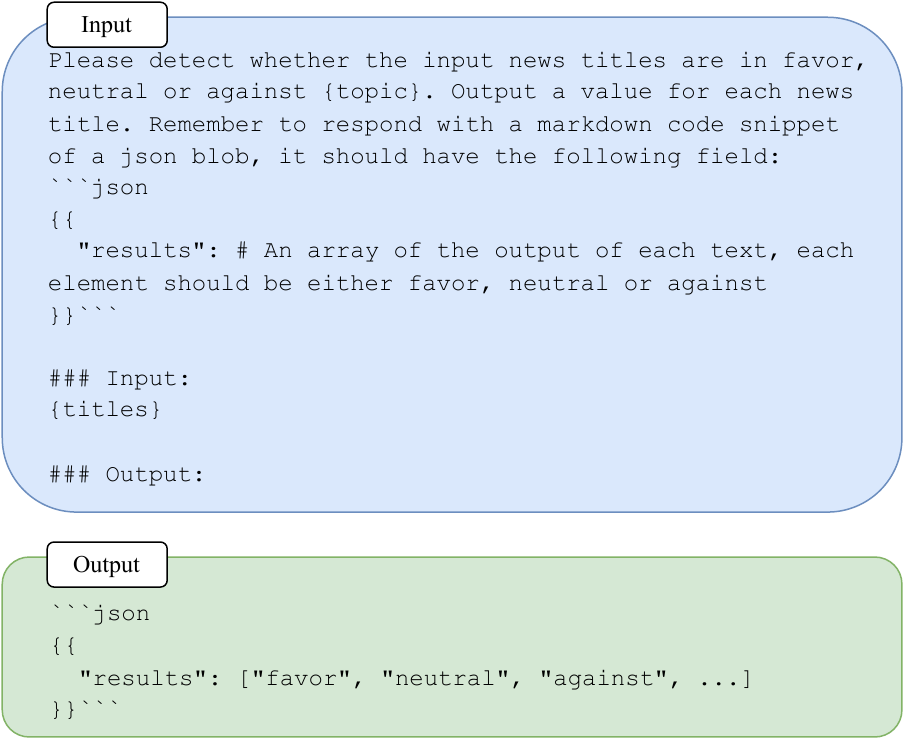}
    \caption{Prompt template for new titles.}
    \label{fig: news_prompt}
\end{subfigure}
\hfill
\begin{subfigure}{0.45\textwidth}
    \includegraphics[width=0.9\textwidth]{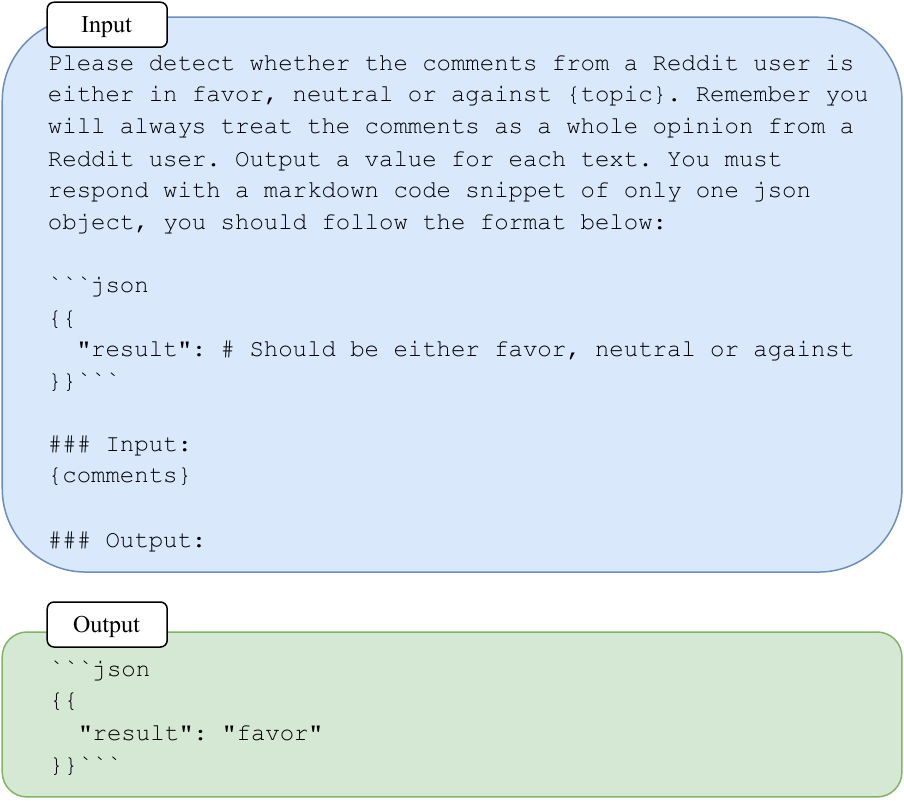}
    \caption{Prompt template for Reddit comments.}
    \label{fig: reddit_prompt}
\end{subfigure}
\caption{Prompt templates for stance classification.}
\end{figure*}

\begin{figure*}[t]
    \centering
    \includegraphics[width=0.9\linewidth]{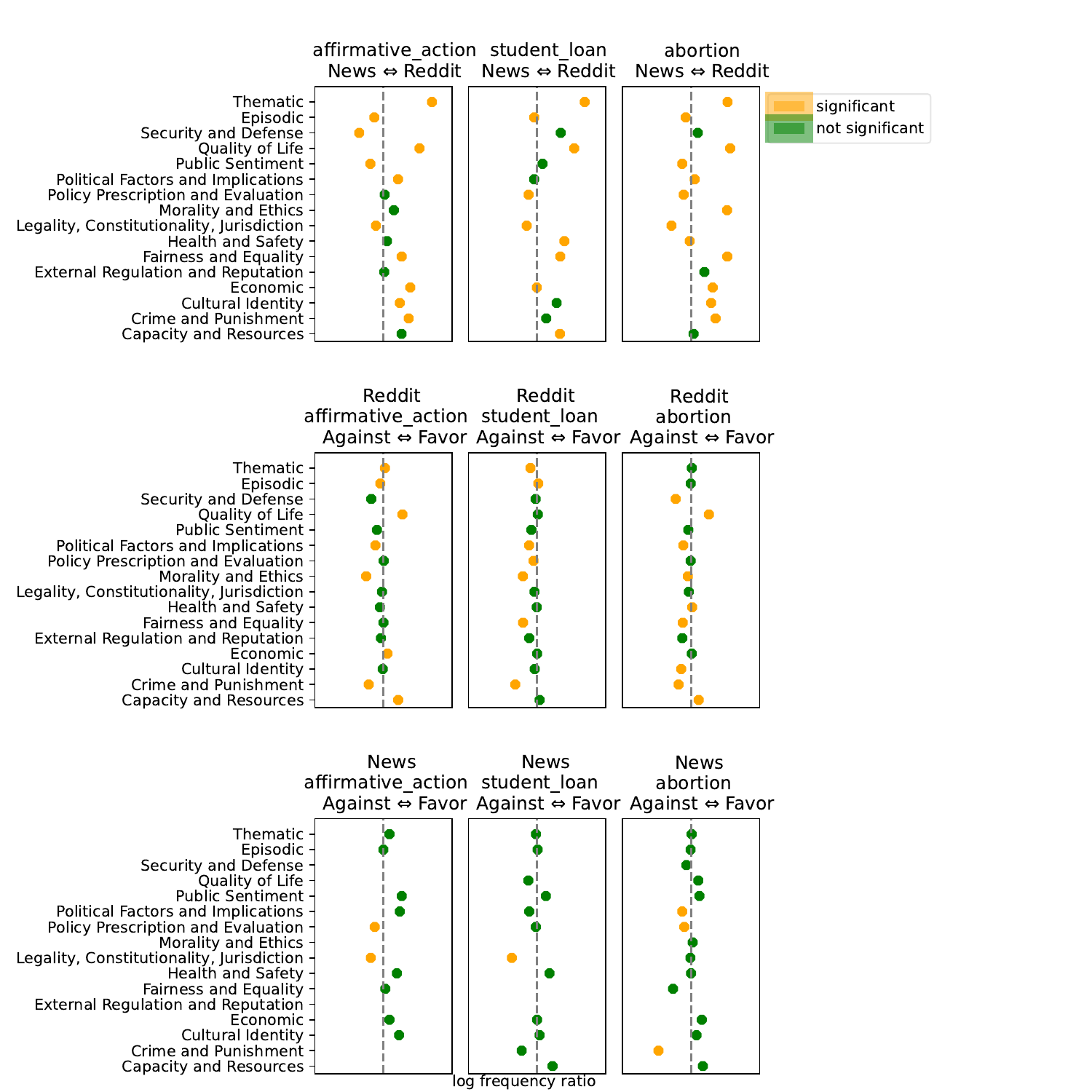}
    \caption{Log frequency ratio of the three topics (\ie affirmative action, student loan, and abortion rights). We perform a z-test to verify if there is a significant difference in the percentage of each frame. If the difference is significant ($p<.05$ after the Bonferroni correction), the dot of the corresponding frame will be colored orange. Otherwise, it will be colored green.}
    \label{fig: log freq ratio}
\end{figure*}

\section{Methods}

\subsection{Data Collection}
Our dataset consists of two components: news titles and Reddit comments. For Reddit comments, we access the archive of Reddit Dumps\footnote{\url{https://github.com/ArthurHeitmann/arctic_shift}} and acquire data that ranges from December 2022 to August 2023. For news titles, we leverage the GoogleNews API\footnote{\url{https://github.com/ranahaani/GNews}} to extract online news titles posted within the same period. The keywords used to collect news titles are shown in Table~\ref{tab:keywords_GNews}.

\subsection{Data Preprocessing}
After obtaining the data, we eliminate duplicated comments and news titles. Subsequently, we employ lemmatization using the Spacy library~\cite{spacy2}. This meticulous process results in a refined collection of 41,143 Reddit comments and 4,784 news titles. An overview of our gathered data is presented in Table~\ref{tab:summary}.

\subsection{Framing Detection}
We employ a readily available multi-label classification model based on the RoBERTa architecture~\cite{mendelsohn2021modeling}. This model efficiently assigns one or multiple frames to each text within the dataset. These frames are categorized into three distinct types: \textit{Issue-Generic Policy}, \textit{Immigration Specific}, and \textit{Narrative}. However, for our study, we exclusively focus on frames originating from the categories of \textit{Issue-Generic Policy} and \textit{Narrative}, as immigration is deemed irrelevant in this context. Descriptions of the selected frames are outlined in Table~\ref{tab:frame}.

\subsection{Stance Classification}
To delve deeper into the distinct framing strategies employed by various groups, we harness the capabilities of ChatGPT ({\ie {\tt gpt-3.5-turbo}})~\cite{ouyang2022training}. Known for its robust zero-shot performance~\cite{qi2023excitements, espejel2023gpt}, ChatGPT enables us to proficiently classify textual content into three distinct categories: \texttt{favor}, \texttt{neutral}, and \texttt{against}. This approach empowers us to identify and categorize the diverse perspectives expressed within the text, thereby enriching our comprehension of framing dynamics. The temperature of ChatGPT is set to 0 for more deterministic responses. We design two different prompt templates for the inference of news titles and Reddit comments. To efficiently utilize each API call of ChatGPT, we infer the new titles in batches, with a batch size of 10. The collection of Reddit comments of an individual user on a certain post is classified as an entirety. The prompt templates are shown in Figures~\ref{fig: news_prompt} and \ref{fig: reddit_prompt}.

\subsection{Model Evaluation}
In order to assess the efficacy of the two chosen models, we sample 60 comments and news titles at random, with each of the three cases contributing 20 instances to the evaluation set. Three researchers conduct independent annotations, and the final labels are assigned based on consensus. Subsequently, we use this dataset to assess the performance of stance classification and framing detection. For classification evaluation, we employ the F1 score as the metric. On the news title side, the F1 score for stance classification reaches 0.70, while for framing detection, it reaches 0.75. On the Reddit comment side, the F1 score for stance classification is 0.55, and for framing detection, it stands at 0.53.

\begin{figure*}[t]
    \centering
    \includegraphics[width=\textwidth]{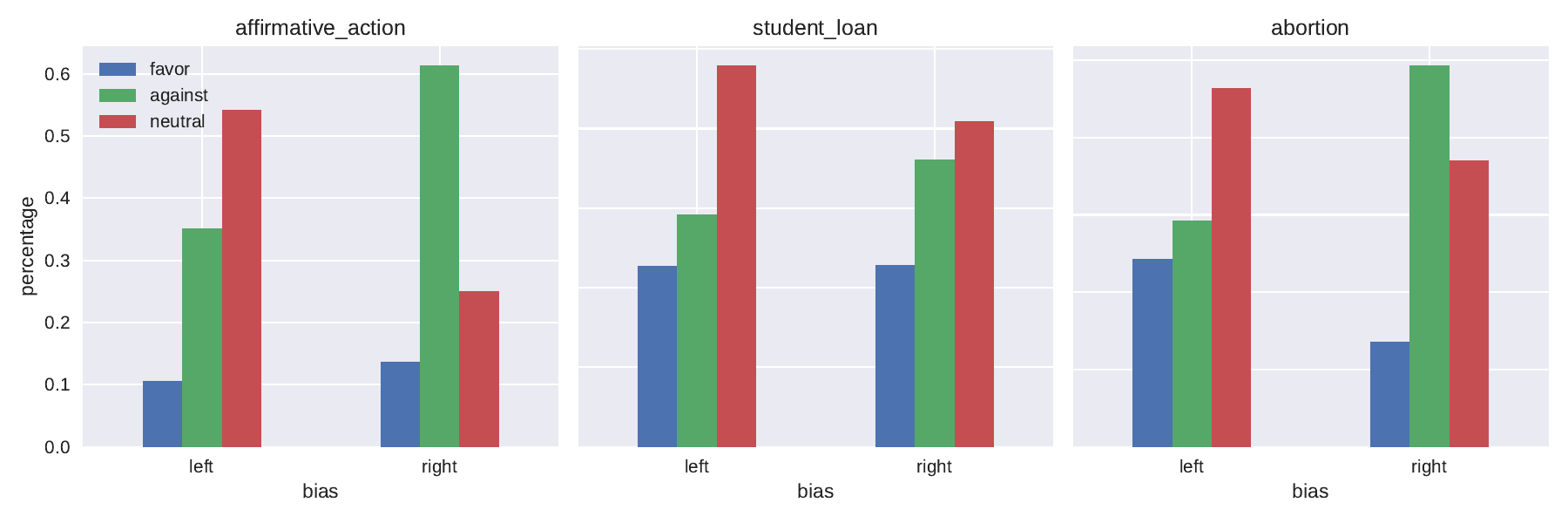}
    \caption{Percentage of \texttt{favor}, \texttt{neutral}, and \texttt{against} titles regarding affirmative action, student loan, and abortion rights of \texttt{Left} and \texttt{Right} media outlets.}
    \label{fig: media1}
\end{figure*}

\section{Results}

\subsection{Cross-platform Comparison}
% In terms of attention to three topics, according to Table \ref{tab:summary}, Abortion  gets the most attention. Additionally, the number of news cover affirmative action takes higher percentages in news than Reddit. 

First, we identify the framing usage by computing the log frequency ratio: 
\begin{equation}
ratio = \frac{n_{fa}/N_A}{n_{fb}/N_B}
\end{equation}
where $n_{fa}$ and $n_{fb}$ are occurrences of a specific frame in group A and group B. $N_A$ and $N_B$ are number of text in target groups, respectively. Here $N_A$ is the number of Reddit comments and $N_B$ is the number of news titles. Additionally, to test whether the log frequency ratio is statistically significant or not, we apply a z-test with multiple comparison corrections. 

The first row in Figure~\ref{fig: log freq ratio} showcases the percentage difference in framing between Reddit comments and news titles. \textbf{Overall, while Reddit and news outlets converge on a few frames, there are still significant variations in several frames across the three issues.} Specifically, Reddit comments primarily exhibit \textit{Thematic} and \textit{Quality of Life} frames across these issues, suggesting that Reddit users often emphasize individual life and experiences and approach discussions in a more abstract manner. The phenomenon that \textit{Fairness and Equality} is more frequent on Reddit implies that Reddit users draw more attention to social justice. In contrast, news outlets predominantly have more frames related to \textit{Episodic}. This underscores the inherent nature of news media reporting, which tends to be more systematic, comprehensive, and sourced compared to discussions on social media. Also, news outlets usually use \textit{Legality, Constitutionality, Jurisdiction} to frame their titles. It is another indicator that media reports are more systematic and comprehensive. 

To identify if there are differences in framing within social media and news outlets, we first represent Reddit comments and news titles using 14-dimensional vectors, with each element of the vector (1 = Yes, 0 = No) corresponding to one of the 14 frames listed in Table~\ref{tab:frame}. Next, we apply the Kmeans++ algorithm~\cite{Arthur2007kmeansTA} to cluster news titles and Reddit comments, respectively. Average inertia is computed to measure the distance for each data point to its assigned centroid, across three cases. For robustness, we trial k=\{5,10,15\} for the number of clusters in Kmeans++. The result is shown in Table~\ref{tab:inertia}. The overall trend is that the average inertia of Reddit comments is larger than that of news titles, suggesting that there is more disagreement in framing on social media than media outlets.

\begin{table}[t]
\caption {Inertia per news title/Reddit comment for the three topics.} 
\label{tab:inertia} 
\centering
\begin{tabular}{l*{3}cr}
\hline
& Affirmative Action & Student Loan & Abortion Rights\\
\hline
Reddit (k=5) &  1.15 & 0.65 & 1.08 \\
News (k=5) & 0.71 &	0.28 & 0.51 \\
Reddit (k=10) &	0.92 &	0.47 & 0.88\\
News (k=10) & 0.47 & 0.15 & 0.36\\
Reddit (k=15) &	0.78 &	0.38 & 0.76\\
News (k=15) & 0.36 & 0.11 & 0.28 \\
% \hline
\hline
\end{tabular}
\end{table}

\begin{table}[t]
\caption{Percentage of stance on affirmative action, student loan, and abortion rights in Reddit comments and news titles.} 
\label{tab:stance summary} 
\centering
\begin{tabular}{l*{3}cr}
\hline
Reddit /\ News & Abortion Rights & Student Loan & Affirmative Action\\
\hline
favor  &  0.27 /\ 0.23  & 0.35 /\ 0.24 & 0.29 /\ 0.08 \\
against  & 0.54 /\ 0.32 &	0.51 /\ 0.27 & 0.65 /\ 0.19 \\
neutral  & 0.19 /\ 0.45 &	0.14 /\ 0.49 & 0.06 /\ 0.72 \\
% \hline
\hline
\end{tabular}
\end{table}

\begin{figure*}[t]
    \centering
    \includegraphics[width=0.9\linewidth]{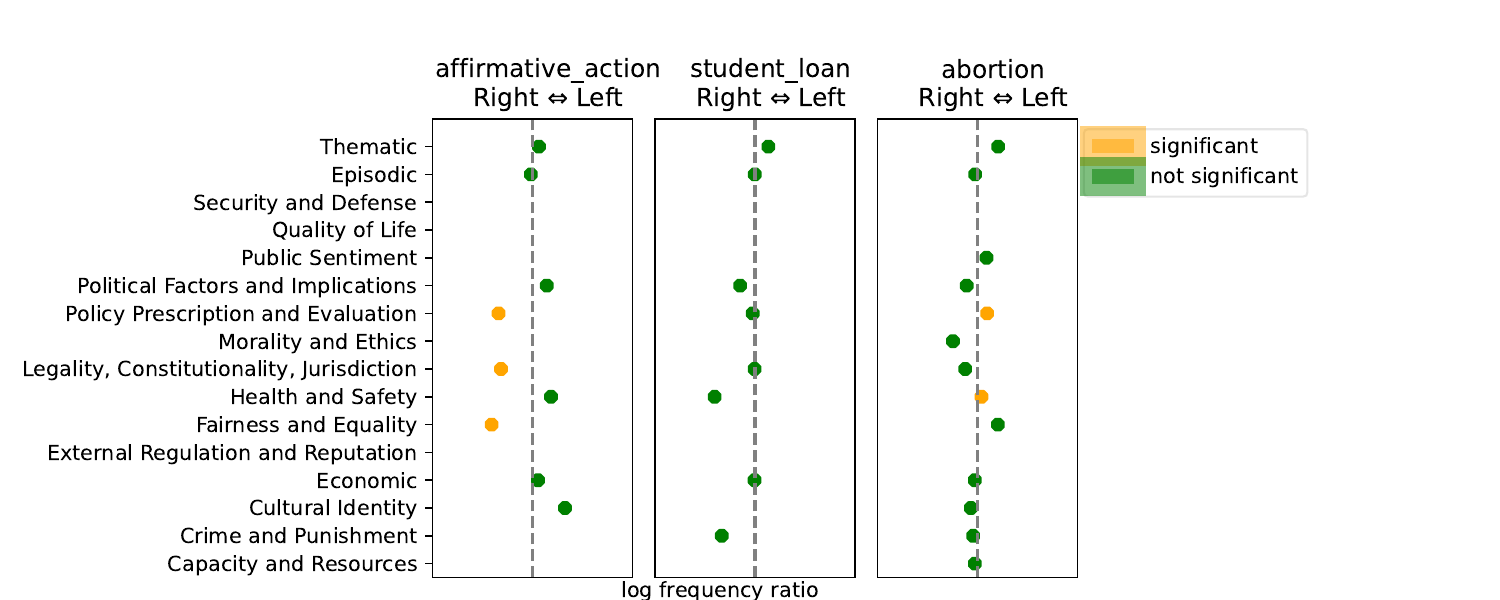}
    \caption{Log frequency ratio of different media groups on three topics (\ie affirmative action, student loan, and abortion rights). We perform a z-test to verify if there is significant difference in the percentage of each frame. If the difference is significant ($p<.05$ after the Bonferroni correction), the dot of the corresponding frame will be colored orange. Otherwise, it will be colored green.}
    \label{fig: media2}
\end{figure*}

\subsection{Stance Analysis}

To investigate whether or not the \textit{supporters} and \textit{opponents} of each issue (\ie affirmative action, student load, abortion rights) frame their comments/titles differently, we compute the log frequency ratio between comments/titles that are classified as \texttt{favor} and \texttt{against}, which is shown in the second and third rows of Figure~\ref{fig: log freq ratio}. Note the frame is left blank in the figure if it is not used by either side (\ie \texttt{favor} or \texttt{against}). 

Table~\ref{tab:stance summary} shows the percentage of comments/titles of \texttt{favor} and \texttt{against}. \textbf{Reddit and news outlets display distinct inclinations in their stances.} Across the three topics, Reddit tends to adopt a predominantly \texttt{against} stance, while news outlets tend to adopt a more \texttt{neutral} position. Notably, the topic with the most pronounced disparity between Reddit and news outlets is affirmative action. This observation suggests that Reddit expresses a higher degree of dissatisfaction with affirmative action, whereas news reports from outlets tend to be less opinionated and adopt a more evasive tone when discussing affirmative action.

In the case of affirmative action discussions on Reddit, the \texttt{against} side often leans towards employing frames related to \textit{Morality and Ethics} and \textit{Crime and Punishment}. Supporters of this perspective view societal issues such as white supremacism as critical concerns in need of resolution. On the \texttt{favor} side, frames such as \textit{Quality of Life} and \textit{Capacity and Resources} are frequently utilized. Those in favor of affirmative action emphasize how access to higher education benefits individuals' quality of life, focusing on the advantages brought by attending college rather than one's racial background. In contrast, news outlets tend to prefer frames such as \textit{Policy Prescription and Evaluation} and \textit{Legality, Constitutionality, Jurisdiction}. For instance, within the \textit{Policy Prescription and Evaluation} frame, media sources list the benefits and significance of affirmative action for educational institutions and higher education in general.

 In the context of student loans, both the \texttt{against} side on Reddit and news outlets exhibit similar patterns to those observed in the discussion of affirmative action. On Reddit, frames such as \textit{Crime and Punishment}, \textit{Morality and Ethics}, and \textit{Fairness and Equality} are commonly employed. Conversely, news outlets predominantly make use of frames related to \textit{Legality, Constitutionality, Jurisdiction}. Much like the discourse surrounding affirmative action, both Reddit and news outlets address the negative aspects of student loans from both micro and macro perspectives. They highlight the micro-level concerns, such as the burden of unaffordable payments on individuals, as well as the macro-level implications, such as the impact of the Supreme Court decisions on Congress.

When it comes to the topic of abortion rights, we observe a shift in the frames used, particularly when compared to the previous topics. On Reddit, the \texttt{against} perspective relies more heavily on frames related to \textit{Security and Defense} and \textit{Cultural Identity}. For instance, within the \textit{Cultural Identity} frame, emphasis is placed on the privilege of women living without the responsibility of a baby, which can be realized through abortion. Conversely, the \texttt{favor} side on Reddit utilizes the \textit{Public Sentiment} frame more frequently, which challenges the notion that abortion should be considered a criminal act. In the case of news outlets taking an \texttt{against} stance on abortion, frames such as \textit{Crime and Punishment}, \textit{Policy Prescription and Evaluation}, and \textit{Political Factors and Implications} are employed to underscore the negative societal consequences of abortion, such as the closure of abortion clinics.

\subsection{Media Outlets Analysis}
We further examine the differences in framing across media outlets with divergent political ideologies. We categorize the media into \texttt{Left} and \texttt{Right}, referring to  Allsides.\footnote{\url{www.allsides.com}} We then extract the frames and stances present in the headlines from each media outlet. 

Figure~\ref{fig: media1} shows the stance of the media on three topics. It is evident that both affirmative action and abortion rights exhibit the most polarized stances. Specifically, while \texttt{Left} media appears largely neutral on affirmative action, the \texttt{Right} media strongly opposes it. A similar trend is observed regarding abortion rights, where \texttt{Left} media leans towards neutrality, while the \texttt{Right} media has a significant oppositional stance. In contrast, opinions on the student loan topic are more convergent across both media outlets, although the \texttt{Right} media displays a slightly stronger opposition.

Figure~\ref{fig: media2} further illustrates the framing differences between \texttt{Left} and \texttt{Right} media. \textbf{The findings reconfirm that affirmative action and abortion rights witness more polarization, while the topic of student loans exhibits greater convergence.} Concerning affirmative action, \texttt{Right} outlets predominantly employ frames such as \textit{Policy Prescription and Evaluation}, \textit{Legality, Constitutionality, Jurisdiction}, and \textit{Fairness and Equality}. This suggests that \texttt{Right} media places a higher emphasis on policy-making and issues of fairness in relation to affirmative action than \texttt{Left} media. Conversely, \texttt{Left} media predominantly uses frames like \textit{Policy Prescription and Evaluation} and \textit{Health and Safety}. These patterns indicate a heightened concern by the \texttt{Left} media about the public health implications of overturning Roe v. Wade in the context of abortion rights.

\section{Discussion and Conclusion}
In this study, we examine disparities in framing across platforms, using Reddit and news headlines related to affirmative action, student loan, and abortion rights as case studies. Our findings indicate that while there is some overlap in framing between social media and traditional media outlets, considerable differences emerge both topic-wise and within specific frame. Compared to traditional news media, social media presents more polarized stances across all frames. Notably, both platforms demonstrate opposition in the \textit{Crime and Punishment} frame concerning abortion rights, potentially reflecting a consensus among the broader public. Finally, we observe significant polarization in the news media's stance on affirmative action and abortion rights, whereas the topic of student loans tends to exhibit greater consensus. These findings \textit{align} with previous research, highlighting the nuanced and substantial divergences in discussions surrounding societal issues within the American political landscape~\cite{lyu2023computational, panbias}.

The differences in framing between traditional and social media platforms hold significant implications for public opinion formation, policy decisions, and the broader political fabric of the US. Framing, integral to the conveyance of messages, offers audiences a particular perspective through which they comprehend multifaceted issues. A disparity in these perspectives between pivotal platforms, such as social media and established news outlets, can result in a disjointed or misperceived public understanding. This segmented perception may, in turn, sway policy-making since policymakers and politicians frequently rely on public sentiment when shaping decisions. Future research could study the varied impacts of these platforms on public opinion, employing diverse methods to extract and analyze framing differences.

\bibliographystyle{IEEEtran}
\bibliography{IEEEfull}

\end{document}